\documentclass[12pt]{article}

\usepackage{sbc-template}
\usepackage{graphicx,url}
\usepackage[utf8]{inputenc}
\usepackage[brazil]{babel}

\title{Aprendizado de máquina aplicado na eletroquímica}

\author{Carlos E. do Egito Araújo\inst{1},  
Lívia F. Sgobbi\inst{2}, 
\\Iwens G. Sene Jr\inst{1}, 
Sergio T. Carvalho\inst{1} 
}

\address{Instituto de Informática -- Universidade Federal de Goiás
  (UFG)\\
  Cep 74690-900 - Goiânia - GO - Brasil -- Brazil
\nextinstitute
  Instituto de Química -- Universidade Federal de Goiás
 (UFG)\\
  Cep 74690-900 - Goiânia - GO - Brasil -- Brazil
  \email{\{cegito,iwens,sergio,livia\_sgobbi\}@ufg.br}
}

\begin{document} 

\maketitle%

\begin{abstract}
This systematic review focuses on analyzing the use of machine learning techniques for identifying and quantifying analytes in various electrochemical applications, presenting the available applications in the literature. Machine learning is a tool that can facilitate the analysis and enhance the understanding of processes involving various analytes. In electrochemical biosensors, it increases the precision of medical diagnostics, improving the identification of biomarkers and pathogens with high reliability. It can be effectively used for the classification of complex chemical products; in environmental monitoring, using low-cost sensors; in portable devices and wearable systems; among others. Currently, the analysis of some analytes is still performed manually, requiring the expertise of a specialist in the field and thus hindering the generalization of results. In light of the advancements in artificial intelligence today, this work proposes to carry out a systematic review of the literature on the applications of artificial intelligence techniques. A set of articles has been identified that address electrochemical problems using machine learning techniques, more specifically, supervised learning.

\end{abstract}
     
\begin{resumo} 
 Esta revisão sistemática foca na análise do emprego de técnicas de aprendizado de máquina para identificar e quantificar analitos em diversas aplicações eletroquímicas, apresentando as aplicações disponíveis na literatura. O aprendizado de máquina é uma ferramenta que pode facilitar a análise e melhoramento da compreensão de processos de diversos analitos. Nos biossensores eletroquímicos, ele eleva a precisão dos diagnósticos médicos, aprimorando a identificação de biomarcadores e patógenos com grande confiabilidade. Pode ser usado na classificação eficaz de produtos químicos complexos; no monitoramento ambiental, usando sensores de baixo custo; em dispositivos portáteis e sistemas vestíveis; dentre outros. Atualmente, a análise de alguns analitos ainda é realizada de forma manual, exigindo o conhecimento de um especialista da área e, dessa forma, dificultando a generalização dos resultados. Em função do avanço na área de inteligência artificial presente hoje, esse trabalho se propõe a realizar uma revisão sistemática da literatura das aplicações de técnicas de inteligência artificial. Foi identificado um conjunto de artigos que resolvem problemas eletroquímicos utilizando técnicas de aprendizado de máquina, mais especificamente, aprendizado supervisionado.

\end{resumo}

\section{Introdução}

A eletroquímica, uma área de pesquisa fundamental, desvenda as interações entre eletricidade e reações químicas. Esta área é vital para o avanço de tecnologias, abrangendo desde o desenvolvimento de baterias até a análise química e síntese eletroquímica, e desempenha um papel importante na detecção e quantificação de analitos em ambientes industriais, biomédicos e de monitoramento ambiental \cite{BardFaulkner2001}.

Paralelamente, o aprendizado de máquina, um campo dinâmico da inteligência artificial, vem ganhando destaque por sua capacidade de processar e interpretar grandes conjuntos de dados. Suas aplicações, que se estendem do reconhecimento de padrões complexos à modelagem preditiva, têm impactado significativamente áreas como a medicina, as finanças e a ciência de materiais \cite{Shehab2022}.

A interseção do aprendizado de máquina com a eletroquímica oferece avanços notáveis, sobretudo na análise eletroquímica de analitos. Esta fusão aprimora a precisão e a eficiência dos métodos eletroquímicos, facilitando a interpretação de sinais eletroquímicos complexos para uma detecção e quantificação mais eficazes de analitos \cite{Zoski2007}. Essa combinação está transformando os métodos analíticos tradicionais e impulsionando o desenvolvimento de novas técnicas e aplicações, desde o monitoramento ambiental até o controle de qualidade em processos industriais \cite{martinez2022aquisiccao, PyzerKnapp2015, UnwinCompton2011}.

Neste trabalho, é explorado como a integração do aprendizado de máquina com a eletroquímica está revolucionando a análise eletroquímica, com um enfoque especial na identificação e quantificação de analitos, abordando os desenvolvimentos recentes e as perspectivas futuras.

Este artigo está assim organizado: a Seção 2 detalha os métodos adotados no planejamento e execução do protocolo do estudo sistemático da literatura; a Seção 3 reporta resultados obtidos ao longo da condução e da análise desses resultados; a Seção 4 apresenta uma síntese dos resultados; e, por fim, a Seção 5 apresenta as considerações finais  e trabalhos futuros.

\section{Métodos}

Esta seção descreve o protocolo utilizado para realizar a Revisão Sistemática da Literatura (RSL), segundo a abordagem proposta por \cite{Nakagawa}. Ele está dividido em três etapas: (I)  Definição  do  protocolo  da  revisão,  constituído  pelas  questões  de  pesquisa,  palavras-chave, sinônimos, estratégia de busca, e os critérios de inclusão e exclusão dos estudos; (II) Condução da revisão, composta pela realização das buscas, seleção dos trabalhos, extração e análise dos dados dos trabalhos selecionados  e,  por fim,  (III) Resultados e Discussão, que apresenta, além dos resultados, as respostas às questões de pesquisa.

A revisão foi projetada para explorar as seguintes questões de pesquisa:
\begin{itemize}
    \item QP1: Como o aprendizado de máquina pode auxiliar no processo de identificação e quantificação de analitos?\\
O aprendizado de máquina é uma ferramenta valiosa na eletroquímica, pois permite desenvolver modelos matemáticos complexos, capazes de analisar grandes volumes de dados experimentais e prever com precisão a identificação e quantificação de analitos.\\
    \item QP2: Quais são as técnicas de aprendizado de máquina mais utilizadas na eletroquímica?\\
As técnicas de aprendizado de máquina mais utilizadas na eletroquímica incluem regressão, classificação, agrupamento e redes neurais artificiais. Essas técnicas permitem abordar diversas tarefas analíticas, como prever concentrações de analitos, categorizar espécies eletroquímicas e identificar padrões nos dados.\\
\end{itemize}

A  {\it string} de busca foi gerada com base no conhecimento dos especialistas das áreas da computação e da química, autores deste artigo:\\
{\it("Electrochemical" OR "electrochemistry") AND ("machine learning") AND ("analysis")}.


A  {\it string} foi utilizada nas seguintes bases bibliográficas: ScienceDirect, IEEE, ACM e Scopus, por serem os principais acervos que se relacionam à área da computação, e a escolha da base Web of Science ocorreu por ser um veículo da área da química. 

Como critério de inclusão, foram selecionados os trabalhos que abordavam ``aprendizado de máquina'' e ``sensores químicos'', como podemos ver no critério CI1: 

\begin{itemize}
   \item[] CI1 – O estudo deve envolver aprendizado de máquina e sensores eletroquímicos. 
\end{itemize}    

Para a exclusão de estudos, foram adotados os seguintes critérios:

\begin{itemize}
     \item[] CE1 - O artigo não descreve um estudo primário.
     \item[] CE2 - O documento retornado não é um artigo.
     \item[] CE3 - O texto completo do artigo não está escrito em inglês.
     \item[] CE4 - O texto completo do artigo não está disponível para acesso.
     \item[] CE5 - O artigo é uma versão mais antiga de outro já considerado.
     \item[] CE6 -  O estudo envolve estudo de baterias.
     \item[] CE7 -  O estudo não envolve aprendizado de máquina e eletroquímica.     
\end{itemize}

Dessa forma, foi possível determinar, a partir da {\it string} de busca desenvolvida e dentre os trabalhos encontrados, quais seriam relevantes para responder às questões de pesquisa.

Como critérios de qualidade, foram utilizados 6 (seis) critérios abaixo listados:

\begin{itemize}
     \item[] CQ1 - O artigo aborda diretamente as questões de pesquisa estabelecidas para a RSL.
     \item[] CQ2 - Os resultados e conclusões apresentados no artigo são baseados em análises rigorosas e evidências robustas.
     \item[] CQ3 - O artigo representa o estado atual do conhecimento no campo.
    \item[] CQ4 - O artigo foi publicado em um periódico ou conferência de reconhecida qualidade e reputação na área de estudo.
     \item[] CQ6 - O artigo contribui com novos conhecimentos ou perspectivas para o campo de estudo.
      \item[] CQ6 - O artigo é bem-escrito, com argumentos claros e estrutura lógica.
\end{itemize}

\section{Resultados}

\begin{figure}[htpb]
\centering
\includegraphics[width=0.8\textwidth]{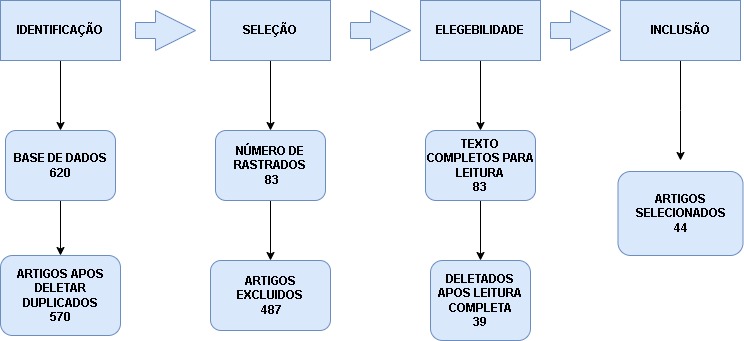}
\caption{Fluxo usado para a seleção final dos artigos.}
\label{fig:figura1}
\end{figure}

Na RSL proposta, foram extraídos 620 artigos das bases de dados citadas anteriormente. Destes, identificou-se 50 como duplicatas. A aplicação rigorosa dos critérios de inclusão e exclusão permitiu a pré-seleção de 83 artigos. Contudo, após uma análise minuciosa do conteúdo e considerando os critérios de qualidade, 39 artigos não se enquadraram completamente nos critérios estabelecidos, resultando em 44 artigos finalmente incluídos na revisão, conforme ilustrado na Figura \ref{fig:figura1}. Na Tabela \ref{tab:selecionados} é apresentada a lista de artigos aceitos para o estudo.

\begin{figure}[htpb]
\centering
\includegraphics[width=0.7\textwidth]{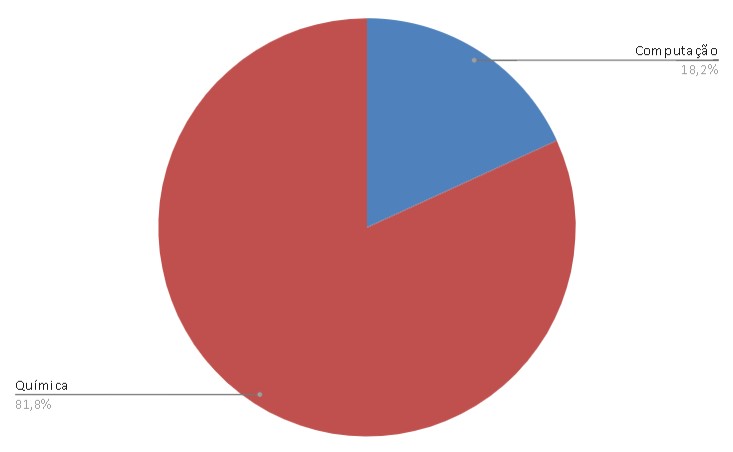}
\caption{Área de publicação dos periódicos.}
\label{fig:figura2}
\end{figure}

Uma análise dos estudos selecionados revelou que 81,8\% das publicações foram veiculadas em periódicos especializados em química, enquanto apenas 18,2\% foram publicadas em periódicos da área de computação, conforme destaca a Figura \ref{fig:figura2}.

\begin{figure}[ht]
\centering
\includegraphics[width=0.8\textwidth]{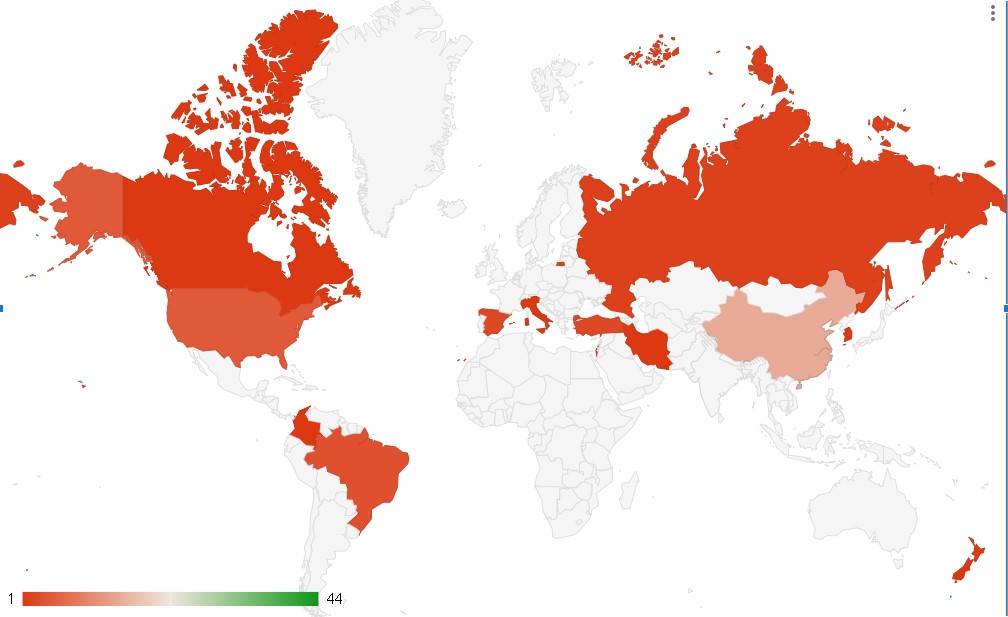}
\caption{Publicação por país.}
\label{fig:figura3}
\end{figure}

Além disso, ao investigar a proveniência destas pesquisas, observou-se que a China lidera em contribuições com 15 publicações. Os Estados Unidos contribuíram com 5 publicações, seguidos pelo Brasil com 4 publicações. Espanha e Turquia tiveram, cada uma, 3 publicações incluídas. As demais publicações possuem origens variadas, conforme detalhado na Figura \ref{fig:figura3}.

\begin{table}[]\tiny\centering
\caption{Artigos selecionados.}
\label{tab:selecionados}
\begin{tabular}{|l|l|l|}
\hline
Id &
  Artigos &
  Autores \\ \hline
S1 &
  \begin{tabular}[c]{@{}l@{}}A Comparison between Different   Machine Learning Approaches Combined with Anodic Stripping Voltammetry for   Copper\\  Ions and pH Detection in Cell Culture Media\end{tabular} & \cite{biscaglia2023comparison} \\ \hline
S2 &
  \begin{tabular}[c]{@{}l@{}}A Machine   Learning-based Multimodal Electrochemical Analytical Device based on   eMoSx-LIG for Multiplexed Detection of Tyrosine\\  and Uric Acid in Sweat and   Saliva\end{tabular} & \cite{ge2022portable} \\ \hline
S3 &
  \begin{tabular}[c]{@{}l@{}}Analyzing the   anodic stripping square wave voltammetry of heavy metal ions via machine   learning: Information beyond\\  a single voltammetric peak\end{tabular} & \cite{ye2020analyzing} \\ \hline
S4 &
  \begin{tabular}[c]{@{}l@{}}A portable smart   detection and electrocatalytic mechanism of mycophenolic acid: A machine   learning-based electrochemical nanosensor to \\ adapt variable-pH silage   microenvironment\end{tabular} & \cite{ge2022portable} \\ \hline
S5 &
  A Portable   Nitrate Biosensing Device Using Electrochemistry and Spectroscopy &
  \cite{asefpour2018portable} \\ \hline
S6 &
  ’All In One’   SARS-CoV-2 variant recognition platform: Machine learning-enabled point of   care diagnostics & \cite{beduk2022all} \\ \hline
S7 &
  An emerging   machine learning strategy for electrochemical sensor and supercapacitor using   carbonized metal–organic framework & \cite{lu2022emerging} \\ \hline
S8 &
  \begin{tabular}[c]{@{}l@{}}Application of   the voltammetric electronic tongue based on nanocomposite modified electrodes   for identifying rice \\ wines of different geographical origins\end{tabular} &
  \cite{wang2019application} \\ \hline
S9 &
  Classification   of As, Pb and Cd Heavy Metal Ions Using Square Wave Voltammetry,   Dimensionality Reduction and Machine Learning &
  \cite{leon2022classification} \\ \hline
S10 &
  \begin{tabular}[c]{@{}l@{}}Detection of   Staphylococcus aureus in milk samples using impedance spectroscopy and data   processing with information visualization \\ techniques and multidimensional   calibration space\end{tabular} &
  \cite{soares2022detection} \\ \hline
S11 &
  Development of   Correction Models for Three-Electrode NO2 Electrochemical Sensor &
  \cite{panjevic2022development} \\ \hline
S12 &
  \begin{tabular}[c]{@{}l@{}}Development of a   simple disposable laser-induced porous graphene flexible electrode for   portable wireless intelligent votammetric\\  nanosensing of salicylic acid in   agro-products\end{tabular} &
  \cite{li2021development} \\ \hline
S13 &
  Deep learning   for pH prediction in water desalination using membrane capacitive   deionization &
  \cite{son2021deep} \\ \hline
S14 &
  \begin{tabular}[c]{@{}l@{}}Electrochemical   and optical detection and machine learning applied to images of genosensors   for diagnosis of prostate cancer with \\ the biomarker PCA3\end{tabular} &
  \cite{rodrigues2021electrochemical} \\ \hline
S15 &
  Electrochemical   Determination of Potassium Ferricyanide using A rtificial Intelligence &
  \cite{asir2019electrochemical} \\ \hline
S16 &
  \begin{tabular}[c]{@{}l@{}}Electrochemical   detection combined with machine learning for intelligent sensing of maleic   hydrazide by using carboxylated PEDOT\\  modified with copper nanoparticles\end{tabular} &
  \cite{sheng2019electrochemical} \\ \hline
S17 &
  Electrochemical   Impedance Spectroscopic Detection of E.coli with Machine Learning &
  \cite{xu2020electrochemical} \\ \hline
S18 &
  Evaluation of   Machine Learning Models on Electrochemical CO2 Reduction Using Human Curated   Datasets &
  \cite{farris2022evaluation} \\ \hline
S19 &
  Fabrication of a   highly sensitive electrochemical sensor for the rapid detection of nimodipine &
  \cite{ma2023fabrication} \\ \hline
S20 &
  \begin{tabular}[c]{@{}l@{}}Green   preparation of amorphous molybdenum sulfide nanocomposite with biochar   microsphere and its voltametric sensing platform for \\ smart analysis of   baicalin\end{tabular} &
  \cite{rao2021green} \\ \hline
S21 &
  \begin{tabular}[c]{@{}l@{}}Green synthesis   of kudzu vine biochar decorated graphene-like MoSe2 with the oxidase-like   activity as intelligent nanozyme sensing \\ platform for hesperetin\end{tabular} &
  \cite{rao2022green} \\ \hline
S22 &
  Highly accurate   heart failure classification using carbon nanotube thin film biosensors and   machine learning assisted data analysis &
  \cite{guo2022highly} \\ \hline
S23 &
  Inflammatory   Stimuli Responsive Non-Faradaic, Ultrasensitive Combinatorial Electrochemical   Urine Biosensor &
  \cite{ganguly2022inflammatory} \\ \hline
S24 &
  Intelligent   Multi-Electrode Arrays as the Next Generation of Electrochemical Biosensors   for Real- Time Analysis of Neurotransmitters &
  \cite{mazafi2018intelligent} \\ \hline
S25 &
  Integration of   an XGBoost model and EIS detection to determine the effect of low inhibitor   concentrations on E. coli &
  \cite{xu2020integration} \\ \hline
S26 &
  IoT and   biosensors: a smart portable potentiostat with advanced cloud-enabled   features &
  \cite{bianchi2021iot} \\ \hline
S27 &
  Low Cost Sensor   With IoT LoRaWAN Connectivity and Machine Learning-Based Calibration for Air   Pollution Monitoring &
  \cite{ali2021low} \\ \hline
S28 &
  Machine Learning   Analysis of Ni/SiC Electrodeposition Using Association Rule Mining and   Artificial Neural Network &
  \cite{kilic2021machine} \\ \hline
S29 &
  \begin{tabular}[c]{@{}l@{}}Machine   learning-assisted Te–CdS@Mn3O4 nano-enzyme induced self-enhanced molecularly   \\ imprinted ratiometric electrochemiluminescence sensor with smartphone for   portable and visual monitoring of 2,4-D\end{tabular} &
  \cite{lu2023machine} \\ \hline
S30 &
  Machine learning   guided electrochemical sensor for passive sweat cortisol detection &
  \cite{shahub2022machine} \\ \hline
S31 &
  Machine-Learning-Guided   Discovery and Optimization of Additives in Preparing Cu Catalysts for CO2   Reduction &
  \cite{guo2021machine} \\ \hline
S32 &
  Machine Learning   Techniques for Chemical Identification Using Cyclic SquareWave Voltammetry &
  \cite{dean2019machine} \\ \hline
S33 &
  Machine learning   to electrochemistry: Analysis of polymers and halide ions in a copper   electrolyte &
  \cite{yoon2021machine} \\ \hline
S34 &
  \begin{tabular}[c]{@{}l@{}}Multilayer   Epitaxial Graphene on Silicon Carbide: A Stable Working Electrode for   Seawater Samples Spiked with Environmental\\  Contaminants\end{tabular} &
  \cite{shriver2020multilayer} \\ \hline
S35 &
  \begin{tabular}[c]{@{}l@{}}MoS2/MWCNTs   porous nanohybrid network with oxidase-like characteristic as electrochemical   nanozyme sensor coupled with machine\\  learning for intelligent analysis of   carbendazim\end{tabular} &
  \cite{zhu2020mos2} \\ \hline
S36 &
  \begin{tabular}[c]{@{}l@{}}Ordinary   microfluidic electrodes combined with bulk nanoprobe produce multidimensional   electric double-layer capacitances towards\\ metal ion recognition\end{tabular} &
  \cite{da2020ordinary} \\ \hline
S37 &
  Phasic dopamine   release identification using convolutional neural network &
  \cite{matsushita2019phasic} \\ \hline
S38 &
  \begin{tabular}[c]{@{}l@{}}Portable   wireless intelligent sensing of ultra-trace phytoregulator $\alpha$-naphthalene   acetic acid using self-assembled  phosphorene/Ti3C2 - \\Mxene nanohybrid with high ambient stability on laser induced porous graphene as nanozyme flexible electrode\end{tabular} &
  \cite{zhu2021portable} \\ \hline
S39 &
  \begin{tabular}[c]{@{}l@{}}Potassium Ferro   cyanide electrochemically detected by Differential Pulse and Square Wave   Voltammetry in a Competition\\  using Gradient Boosting as Machine Learning   Algorithm\end{tabular} &
  \cite{shama2020potassium} \\ \hline
S40 &
  \begin{tabular}[c]{@{}l@{}}Resolution of   opiate illicit drugs signals in the presence of some cutting agents with use   of a voltammetric sensor array and machine\\  learning strategies\end{tabular} &
  \cite{ortiz2022resolution} \\ \hline
S41 &
  Smart   electrochemical sensing of xylitol using a combined machine learning and   simulation approach &
  \cite{uwaya2022smart} \\ \hline
S42 &
  Soft Hydrogel   Actuator for Fast Machine-Learning-Assisted Bacteria Detection &
  \cite{lavrentev2022soft} \\ \hline
S43 &
  Time-Lapse   Electrochemical Impedance Detection of Bacteria Proliferation for Accurate   Antibiotic Evaluation &
  \cite{chen2022time} \\ \hline
S44 &
  \begin{tabular}[c]{@{}l@{}}Voltammetric   sensing using an array of modified SPCE coupled with machine learning   strategies for the improved \\ identification of opioids  in presence of cutting   agents\end{tabular} &
  \cite{ortiz2021voltammetric} \\ \hline
\end{tabular}
\end{table}

\section{Discussão}

Nesta seção, são discutidas a análise e a síntese das informações coletadas dos 44 estudos selecionados. Os resultados dessa análise e síntese são fundamentais para responder às questões de pesquisa propostas nesta RSL, conforme detalhado a seguir.

QP1 - Como aprendizado de máquina pode auxiliar no processo de identificação e quantificação de analitos?

\begin{figure}[ht]
\centering
\includegraphics[width=0.7\textwidth]{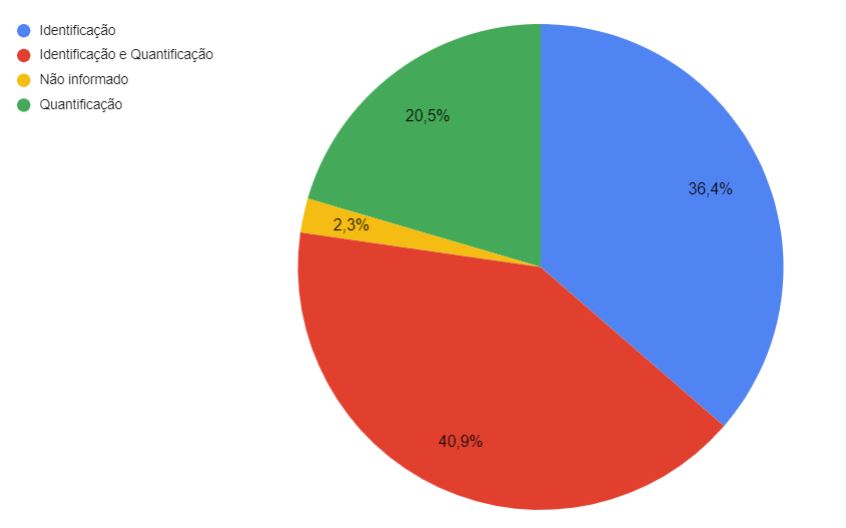}
\caption{Categorias dos artigos selecionados.}
\label{fig:figura4}
\end{figure}

De acordo com os dados apresentados na Figura \ref{fig:figura4}, é possível observar diferentes abordagens nos artigos revisados. A maioria dos estudos (40,9\%) concentra-se na identificação e quantificação dos analitos, indicando uma forte ênfase na obtenção de informações abrangentes sobre as espécies químicas em análise. Em contrapartida, aproximadamente 36,4\% dos artigos focam exclusivamente na identificação dos analitos, demonstrando uma preocupação em caracterizar as substâncias presentes em sistemas eletroquímicos.

Além disso, cerca de 20,5\% dos artigos dedicam-se especificamente à quantificação dos analitos, evidenciando o interesse em determinar com precisão as concentrações das substâncias de interesse. É relevante notar que 2,3\% dos estudos não abordam nem a identificação nem a quantificação dos analitos, mas concentram-se na integração de técnicas de aprendizado de máquina e eletroquímica. Essa abordagem indica a busca por soluções inovadoras que podem não se concentrar diretamente na análise química tradicional, mas que aproveitam os benefícios da inteligência artificial na interpretação de dados eletroquímicos de maneira mais ampla. Essa diversidade de abordagens reflete a amplitude de possibilidades que o campo da eletroquímica e do aprendizado de máquina oferece para a análise de analitos, permitindo a escolha da estratégia mais adequada para cada cenário de pesquisa.

Com base nas conclusões dos estudos mencionados, o aprendizado de máquina demonstrou ser uma ferramenta valiosa no processo de identificação e quantificação de analitos na eletroquímica. Diversas técnicas de aprendizado de máquina foram aplicadas com sucesso em diferentes contextos eletroquímicos, contribuindo para avanços significativos em áreas como a detecção de poluentes, o monitoramento ambiental, o diagnóstico médico e a análise de substâncias específicas. Essas técnicas têm a capacidade de lidar com a complexidade dos dados obtidos em experimentos eletroquímicos, melhorando a precisão e eficiência das análises.

Os estudos revisados destacaram várias técnicas de aprendizado de máquina frequentemente utilizadas na eletroquímica. Redes Neurais Artificiais (ANNs), por exemplo, foram aplicadas com êxito em várias pesquisas, como no estudo de \cite{ge2022portable}, que utilizou ANNs para modelar a relação entre o pH das amostras de silagem de grama e a concentração de ácido micofenólico, contribuindo para a análise de qualidade de alimentos na eletroquímica. Além disso, \cite{lu2022emerging} destacaram a aplicação de ANNs na análise da concentração de niclosamida em soluções aquosas.

Outras técnicas populares incluem algoritmos de aprendizado de máquina, como Support Vector Machines (SVM), Naive Bayes, Random Forest, Árvores de Decisão, Redes Neurais Convolucionais (CNNs) e Redes Neurais Profundas (DNNs). Por exemplo, \cite{biscaglia2023comparison} construíram um modelo de previsão para medir a concentração de íons de cobre em meios comerciais de cultura de células, utilizando Árvore de Decisão, SVM, Naive Bayes e Random Forest. \cite{ye2020analyzing} enfatizaram o uso do SVM na detecção de íons de metais pesados enquanto  \cite{leon2022classification}  utilizaram $k$-vizinhos mais próximos, Naive Bayes e Árvores de Decisão para classificar íons de metais pesados.

Além disso, \cite{asefpour2018portable} empregaram diversos algoritmos, incluindo SVM, para prever a concentração de nitrato em amostras líquidas. \cite{tanak2022multiplexed}  aplicaram CNNs para melhorar a detecção de eventos específicos em contextos eletroquímicos, enquanto \cite{beduk2022all} usaram DNNs para identificar variantes do vírus SARS-CoV-2 em amostras de esfregaço nasofaríngeo.

Essas técnicas, quando aplicadas adequadamente, proporcionam melhorias significativas na identificação e quantificação de analitos na eletroquímica, demonstrando a diversidade de abordagens disponíveis para resolver problemas específicos nesse campo. Portanto, o aprendizado de máquina se torna uma ferramenta versátil e poderosa para os cientistas e pesquisadores da eletroquímica, permitindo abordagens personalizadas de acordo com as necessidades de cada estudo citado anteriormente. 

QP2 - Quais são as técnicas de aprendizado de máquina mais utilizadas na eletroquímica?

\begin{figure}[ht]
\centering
\includegraphics[width=1\textwidth]{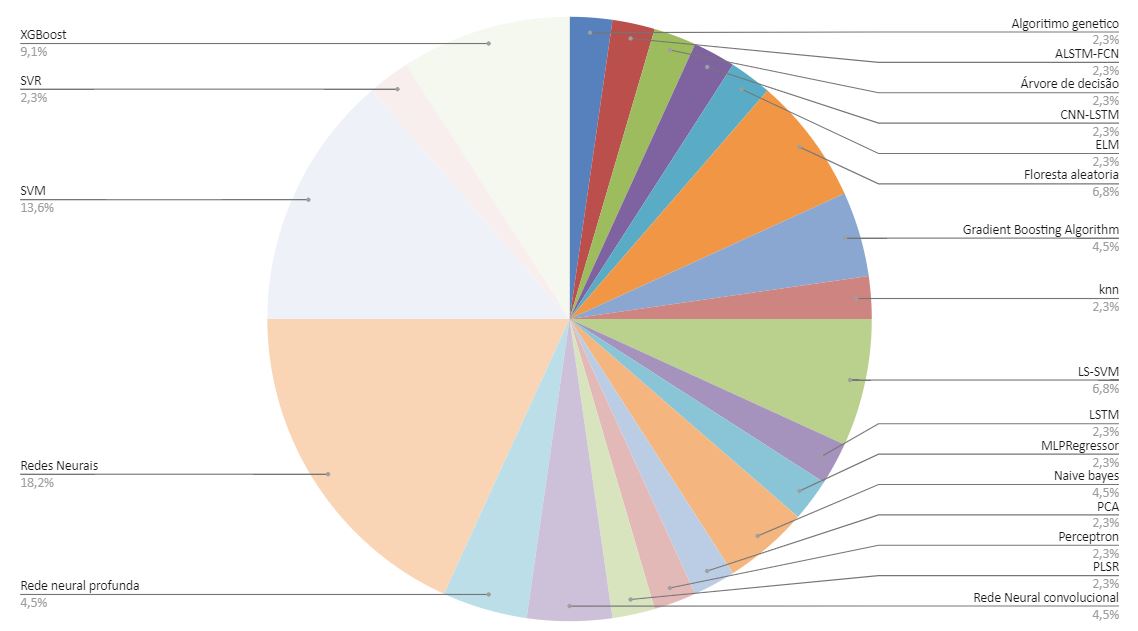}
\caption{Técnicas empregadas nos estudos.}
\label{fig:figura5}
\end{figure}

Na Figura \ref{fig:figura5}, é possível observar a distribuição das abordagens utilizadas. Nota-se que a utilização de Redes Neurais Artificiais (ANNs) representa aproximadamente 18,2\% do total, seguida pela Máquina de Vetores de Suporte (SVM), que corresponde a cerca de 13,6\% das escolhas. Em terceiro lugar, encontra-se o XGBoost com 9,1\%, enquanto a quarta e a quinta posição são ocupadas pelo Random Forest e LS-SVM, ambas com 6,8\% de representação.

Além dessas técnicas, outras abordagens também foram exploradas nos estudos, como o Gradient Boosting Algorithm, o Naive Bayes, a Rede Neural Convolucional e a Rede Neural Profunda, que têm uma parcela de 4,5\% cada nas escolhas dos pesquisadores. O restante dos algoritmos utilizados tem uma representatividade de 2,3\% cada.

Este panorama revela a diversidade de abordagens adotadas na pesquisa, demonstrando que diferentes técnicas de aprendizado de máquina são exploradas em proporções variadas para abordar os desafios específicos de cada estudo. Essa variedade de opções destaca a importância da seleção adequada de algoritmos de acordo com o contexto e os objetivos de cada pesquisa.

Em vários estudos mencionados, incluindo os conduzidos por \cite{ge2022portable},  \cite{lu2022emerging}, \cite{panjevic2022development}, \cite{sheng2019electrochemical},\cite{ali2021low}, \cite{zhu2020mos2}, \cite{zhu2021portable} e \cite{uwaya2022smart}, as Redes Neurais Artificiais (ANNs) desempenharam um papel central na análise e previsão de dados em contextos diversos.

No estudo conduzido por \cite{ge2022portable}, o diferencial está na aplicação de ANNs para modelar a relação entre o pH das amostras de silagem de grama e a concentração de ácido micofenólico, destacando a capacidade das ANNs de prever concentrações de compostos em amostras de pH variável, essenciais na eletroquímica relacionada à qualidade de alimentos. Em \cite{lu2022emerging}, o destaque é a aplicação da ANNs para análise da concentração de niclosamida em soluções aquosas, contribuindo para a eletroquímica em análises químicas, nas quais a ANN pode ser valiosa na detecção de substâncias químicas específicas em soluções. O estudo de  \cite{panjevic2022development} diferencia-se ao aplicar algoritmos de ANN para corrigir dados de sensores de qualidade do ar de baixo custo, especialmente sensores de N$O_2$, oferecendo uma contribuição significativa para a eletroquímica aplicada à monitoração ambiental de forma acessível. Em \cite{sheng2019electrochemical}, o diferencial reside na combinação de detecção eletroquímica com aprendizado de máquina para melhorar a detecção de hidrazida maleica, destacando a ANN como uma ferramenta poderosa na otimização da análise de dados eletroquímicos. O estudo de \cite{ali2021low}destaca-se por calibrar sensores de monóxido de carbono (CO) usando diferentes algoritmos, incluindo ANN, para melhorar a precisão da detecção, contribuindo para a eletroquímica aplicada à monitoração ambiental e segurança. Em \cite{zhu2020mos2}, o diferencial está na aplicação de ANNs para análise inteligente de resíduos de carbendazim, evidenciando a capacidade das ANNs em lidar com dados complexos relacionados à detecção de resíduos químicos em alimentos, relevante para a segurança alimentar na eletroquímica. Já \cite{zhu2021portable} contribui para a eletroquímica ao utilizar ANNs para analisar a concentração de ácido $\alpha$-naftalenoacético (NAA) em amostras agrícolas, destacando a aplicação de ANNs na agricultura e segurança alimentar. Em \cite{uwaya2022smart}, o diferencial está na aplicação de ANNs para relacionar dados eletroquímicos com a concentração de xilitol em goma sem açúcar, oferecendo uma contribuição para a indústria alimentar por meio da análise inteligente de dados eletroquímicos.

Todos esses estudos frequentemente envolveram a coleta e preparação de amostras de dados pelos próprios pesquisadores, em vez de recorrer a bases de dados externas.
A coleta e preparação de amostras pelos pesquisadores acrescentam um nível adicional de confiabilidade e controle aos estudos, garantindo que os dados sejam relevantes e específicos para os objetivos de pesquisa. Essa abordagem personalizada destaca o comprometimento dos pesquisadores com a qualidade dos dados usados em suas análises.
A escolha da ANN foi justificada em muitos casos devido à sua capacidade de lidar com complexidades intrínsecas aos dados e fornecer previsões precisas com base em informações de entrada, destacando seu papel crucial na análise inteligente de dados em uma variedade de aplicações, desde a detecção de resíduos químicos até o monitoramento de qualidade do ar e o desenvolvimento de sensores de baixo custo.

A escolha de ANNs é motivada por sua capacidade de lidar com dados complexos e não lineares, o que é comum em muitos contextos da eletroquímica. Além disso, as ANNs pode oferecer previsões precisas, o que é essencial em aplicações como a detecção de resíduos químicos, monitoramento ambiental e desenvolvimento de sensores, em que a precisão é fundamental para resultados confiáveis e eficazes. Isso destaca a importância das ANNs como uma ferramenta versátil na análise de dados eletroquímicos em diversos campos de pesquisa.

Vários estudos relacionados à eletroquímica têm utilizado algoritmos de aprendizado de máquina para resolver problemas complexos. No artigo de \cite{biscaglia2023comparison}, os pesquisadores construíram um modelo de previsão para medir a concentração de íons de cobre em meios comerciais de cultura de células. Eles usaram quatro algoritmos diferentes de aprendizado de máquina, como Árvore de Decisão, Máquina de Vetores de Suporte (SVM), Naive Bayes e Random Forest, para melhorar a precisão da detecção destes íons. O SVM destacou-se com a melhor precisão de 96,6\%. \cite{ye2020analyzing} trabalhou com a detecção de íons de metais pesados usando análise eletroquímica e aprendizado de máquina, com ênfase no SVM, que alcançou mais de 90\% de precisão. Já \cite{leon2022classification} utilizaram algoritmos como $k$-vizinhos mais próximos, Naive Bayes, Árvores de Decisão, entre outros, para classificar íons de metais pesados. Mais uma vez, o SVM obteve alta precisão, comprovando sua eficácia como o algoritmo de escolha. \cite{asefpour2018portable} também aplicaram diversos algoritmos, incluindo SVM, para prever a concentração de nitrato em amostras líquidas, e o SVM apresentou o melhor desempenho.

Em todos esses estudos, os pesquisadores escolheram algoritmos de aprendizado de máquina devido à sua eficácia em lidar com dados complexos e não lineares. Eles também aplicaram técnicas de análise de dados eletroquímicos em várias aplicações, desde a detecção de poluentes até o diagnóstico médico, destacando o potencial do aprendizado de máquina na eletroquímica. De maneira notável, todos esses estudos chegaram à conclusão de que o SVM foi o algoritmo com os melhores resultados, fornecendo evidências sólidas de sua superioridade em contextos eletroquímicos. Embora os detalhes sobre a origem dos dados variem, em muitos casos, os próprios pesquisadores prepararam as amostras, demonstrando o comprometimento com a qualidade da pesquis. Esses estudos mostram como algoritmos de aprendizado de máquina, com ênfase no SVM, são ferramentas valiosas na análise de dados complexos em diferentes contextos eletroquímicos.

Em diversos estudos mencionados, incluindo os realizados por \cite{ma2023fabrication}, \cite{xu2020electrochemical}, \cite{shahub2022machine} e \cite{chen2022time}, a aplicação de algoritmos de aprendizado de máquina tem se destacado como uma ferramenta importante na análise de dados eletroquímicos. Esses estudos compartilham a utilização de algoritmos de aprendizado de máquina, notavelmente o Extreme Gradient Boosting (XGBoost), e demonstram resultados promissores em suas respectivas áreas de pesquisa.

O XGBoost, em particular, tem sido amplamente escolhido devido à sua eficácia, capacidade de lidar com múltiplos parâmetros e aplicabilidade em diversas áreas. 
Em resumo, esses estudos destacam o papel crucial do algoritmo XGBoost na análise de dados eletroquímicos, ilustrando sua capacidade de melhorar a precisão e eficácia em uma variedade de aplicações. 

É importante ressaltar que, em todos esses estudos, os autores não utilizaram uma base de dados preexistente, mas sim prepararam suas próprias amostras. Por exemplo, \cite{ma2023fabrication} trituraram 20 comprimidos de nimodipina em pó e dissolveram-nos em etanol para criar suas amostras. \cite{xu2020integration} obtiveram amostras experimentais por meio de espectroscopia de impedância eletroquímica, preparando-as para treinar e testar seus modelos de aprendizado de máquina. Da mesma forma, \cite{shahub2022machine} desenvolveram seu próprio conjunto de dados para avaliar o desempenho de catalisadores, enquanto \cite{chen2022time} utilizaram amostras relacionadas à detecção de impedância eletroquímica.

Os estudos conduzidos por \cite{soares2022detection}, \cite{farris2022evaluation} e \cite{ganguly2022inflammatory} ilustram a crescente relevância e eficácia do uso de algoritmos de aprendizado de máquina na análise de dados eletroquímicos. Essas pesquisas compartilham a escolha de algoritmos como Árvore de Decisão e, notavelmente, o Random Forest no caso de \cite{ganguly2022inflammatory}. É possível observar que o estudo de \cite{ganguly2022inflammatory} obteve os resultados mais relevantes, com uma precisão máxima de aproximadamente 98,4\% na classificação de amostras de urina relacionadas a doenças inflamatórias. Esses estudos demonstram o potencial do aprendizado de máquina para melhorar a precisão e a robustez das análises eletroquímicas, abrindo caminho para aplicações práticas em diversos campos, desde a detecção de bactérias em amostras de alimentos até a classificação de estados de doenças inflamatórias com base em dados de urina. 


Os estudos realizados por \cite{li2021development}, \cite{rao2021green} e \cite{rao2022green} compartilham uma abordagem comum na aplicação de algoritmos de aprendizado de máquina para análise de dados eletroquímicos, com foco na relação entre os dados dos sensores e as concentrações de substâncias específicas, como ácido salicílico, baicalina e hesperetina. Os algoritmos escolhidos incluem ANNs, Máquinas de Vetores de Suporte com Mínimos Quadrados (LSSVM) e Rede Neural Artificial Baseada em Probabilidade (PB-ANN). No entanto, a quantidade exata de amostras utilizadas nos estudos não foi especificada e as amostras foram preparadas pelos próprios autores.

Quanto aos resultados, o estudo de \cite{li2021development} concluiu que o algoritmo LSSVM superou outros modelos de aprendizado de máquina na análise quantitativa do ácido salicílico. Em contraste, o estudo de \cite{rao2021green} comparou modelos de Regressão Linear Simples, ANN e LSSVM para a identificação e quantificação da baicalina, e embora a ANN tenha obtido um desempenho ligeiramente superior, o LSSVM foi considerado mais adequado para aplicações práticas. Por fim, o estudo de \cite{rao2022green} destacou o desempenho superior do algoritmo LS-SVM em relação ao PB-ANN na análise da hesperetina, com base em métricas como Relação de Desempenho de Diferença (RPD), Erro Médio Quadrático de Raiz Padrão (RMSEP) e Erro Médio Absoluto de Predição (MAEP). 
O desempenho superior do algoritmo LSSVM em dois dos estudos sugere que ele pode ser uma escolha particularmente eficaz para análises eletroquímicas.

Os estudos apresentados por \cite{asir2019electrochemical} e \cite{shama2020potassium} compartilham uma abordagem comum na aplicação do algoritmo de Gradient Boosting como técnica de aprendizado de máquina para a análise de dados eletroquímicos, especificamente voltamogramas, visando classificação e quantificação de analitos e interferentes em variantes de ferricianeto de potássio.
O algoritmo Gradient Boosting foi escolhido em ambos os estudos devido à sua aplicabilidade em tarefas de classificação e previsão em várias áreas, incluindo eletroquímica, e ao seu desempenho comprovado em estudos anteriores. No entanto, é importante observar que o estudo de \cite{asir2019electrochemical} obteve uma taxa de classificação de 76,6\% ao considerar 30\% das variantes, mas essa taxa diminuiu para 60\% ao aumentar a porcentagem de variantes consideradas para 40\%. Por outro lado, o estudo de \cite{shama2020potassium} alcançou uma precisão de 75\% para a técnica de Voltametria de Pulso Diferencial (DPV) e 60\% para a Voltametria de Onda Quadrada (SWV), destacando a importância do algoritmo de Gradient Boosting na classificação e quantificação de dados eletroquímicos complexos.

No que diz respeito aos dados de teste, o estudo de \cite{asir2019electrochemical} utilizou um conjunto de dados composto por 100 variantes de ferricianeto de potássio, mas não especificou a origem ou preparação das amostras, se foram obtidas por meio de preparação ou retiradas de um banco de dados existente. Já o estudo de \cite{shama2020potassium} não forneceu detalhes sobre a quantidade ou origem das amostras utilizadas.
Enfim, esses estudos evidenciam que a aplicação do algoritmo de Gradient Boosting representa uma abordagem eficaz na análise de dados eletroquímicos complexos, permitindo a classificação e quantificação de analitos e interferentes em voltamogramas. A precisão alcançada varia entre os estudos, mas destaca a viabilidade dessa abordagem para aplicações eletroquímicas, ressaltando a importância do algoritmo de Gradient Boosting nesse contexto.

Os estudos conduzidos por \cite{tanak2022multiplexed} e \cite{ortiz2021voltammetric} compartilham uma abordagem comum na aplicação de vários algoritmos de aprendizado de máquina, incluindo Random Forest, Naive Bayes e SVM, para identificar drogas de abuso e agentes de corte em amostras analisadas, com ênfase na análise voltamétrica. Ambos os estudos observaram que tanto o Random Forest quanto o Naive Bayes produziram resultados satisfatórios em todos os indicadores avaliados, enquanto o SVM demonstrou algum grau de classificação incorreta em amostras específicas. No entanto, nenhum dos artigos menciona explicitamente qual dos algoritmos apresentou o melhor desempenho na identificação das substâncias em questão.
A escolha desses algoritmos foi baseada em sua aplicação em quimiometria e em estudos semelhantes. 
Os estudos analisaram um total de 20 amostras, representando a determinação de várias substâncias, com ênfase em opioides e agentes de corte. Embora os artigos não especifiquem a origem das amostras, sugerem que eles próprios podem ter sido responsáveis pela sua preparação.

Nos estudos de \cite{lu2023machine} e \cite{matsushita2019phasic} as CNNs são escolhidas como algoritmos devido à sua eficácia no processamento de imagens e reconhecimento de padrões. No entanto, \cite{lu2023machine} utiliza a combinação de CNNs com uma rede Siamesa, enquanto \cite{matsushita2019phasic} se concentra na aplicação direta de CNNs. Nenhum dos artigos fornece uma avaliação quantitativa detalhada do desempenho das CNNs, como métricas de precisão ou outros indicadores específicos de desempenho. Além disso, ambos os textos não esclarecem a origem das amostras de imagens utilizadas nos experimentos, não especificando se foram coletadas especificamente para o estudo ou se provêm de um banco de dados preexistente. 

Os estudos mencionados por \cite{beduk2022all} compartilham a utilização do mesmo algoritmo de aprendizado de máquina, a Rede Neural Profunda (DNN), para diferentes aplicações na área da saúde.
No primeiro estudo, a DNN foi empregada para a identificação de variantes do vírus SARS-CoV-2 em amostras de esfregaço nasofaríngeo. Os resultados foram impressionantes, com taxas de acurácia variando de 98,7\% a 100\%, dependendo da variante em questão. No segundo estudo, a DNN foi aplicada para prever os níveis de peptídeo natriurético tipo B (BNP) em amostras de sangue de pacientes com insuficiência cardíaca, alcançando acurácia de 95,6\% a 97,2\%. 
Oos autores não conduziram a preparação das amostras de esfregaço nasofaríngeo ou de sangue por conta própria, obtendo essas amostras de laboratórios hospitalares ou de pacientes.
Em relação aos resultados, ambos os estudos demonstram que a DNN possui um notável potencial em aplicações clínicas, melhorando a precisão de diagnósticos e análises de dados complexos. No entanto, no segundo estudo, é mencionada uma limitação relacionada ao tamanho da amostra e à distribuição desigual entre as classes, o que afetou a estabilidade do desempenho diagnóstico da DNN. 

Além dos estudos previamente mencionados, é relevante ressaltar que há mais 12 estudos que relatam o uso de aprendizado de máquina em eletroquímica: \cite{kammarchedu2022machine}, \cite{wang2019application}, \cite{son2021deep}, \cite{xu2020electrochemical}, \cite{mazafi2018intelligent}, \cite{lu2023machine}, \cite{dean2019machine}, \cite{yoon2021machine}, \cite{shriver2020multilayer}, \cite{da2020ordinary}, \cite{ortiz2022resolution}, e \cite{lavrentev2022soft}.
Embora tenham havido variações nas aplicações e na escolha de algoritmos específicos, há notáveis pontos de convergência entre esses estudos, por exemplo muitos desses estudos optaram por preparar suas próprias amostras de dados, em vez de depender de bases de dados externas. 
No que diz respeito aos algoritmos de aprendizado de máquina escolhidos, eles destacam a eficácia de algoritmos específicos, tais como ANN, XGBoost, Gradient Boosting, Random Forest, Naive Bayes, SVM e Redes Neurais Convolucionais (CNN). A escolha de cada algoritmo é fundamentada na sua adequação às tarefas específicas de análise e previsão de dados eletroquímicos.
É importante notar que, embora os resultados possam variar entre os estudos, em muitos casos, os algoritmos de aprendizado de máquina demonstraram melhorar significativamente a precisão, eficiência e robustez das análises eletroquímicas. 


A análise de todos os estudos revisados revela uma notável diversidade de abordagens no campo da eletroquímica e aprendizado de máquina. A maioria dos estudos concentra-se na identificação e quantificação dos analitos, refletindo uma ênfase em obter informações detalhadas sobre as espécies químicas em análise. Esse enfoque é de importância crítica para a compreensão da composição química dos sistemas eletroquímicos.

Além disso, aproximadamente 36,4\% dos artigos concentram-se exclusivamente na identificação dos analitos, evidenciando uma preocupação com a caracterização das substâncias presentes em sistemas eletroquímicos, mesmo sem ênfase na quantificação. Essa abordagem pode ser relevante para determinar a presença ou ausência de substâncias específicas em uma amostra.

Cerca de 20,5\% dos estudos se dedicam especificamente à quantificação dos analitos, refletindo o interesse em determinar com precisão as concentrações das substâncias de interesse. Isso é fundamental em aplicações que exigem medições quantitativas, como monitoramento ambiental ou análise clínica.

É importante notar que 2,3\% dos estudos não abordam nem a identificação nem a quantificação dos analitos, mas, em vez disso, concentram-se na integração de técnicas de aprendizado de máquina e eletroquímica. Essa abordagem aponta para a busca de soluções que não se prendem estritamente à análise química tradicional, aproveitando os benefícios da inteligência artificial na interpretação de dados eletroquímicos de forma mais abrangente.


Em resumo, os estudos revisados indicam que o aprendizado de máquina é uma ferramenta valiosa para a identificação e quantificação de analitos na eletroquímica. Diversas técnicas de aprendizado de máquina, como ANNs, SVM, Naive Bayes, Random Forest dentre outras, têm demonstrado sucesso em várias aplicações eletroquímicas, tais como detecção de poluentes, monitoramento ambiental, diagnóstico médico e análise de substâncias específicas.


É importante destacar que a preparação e coleta de amostras pelos próprios pesquisadores agregam um nível adicional de confiabilidade e controle às investigações. Isso garante que os dados sejam relevantes e específicos para os objetivos de pesquisa, um fator crucial para a validade das análises.


\section{Considerações finais}

Os estudos revisados abordam um amplo espectro de aplicações que vão desde a detecção de poluentes até o diagnóstico médico, demonstrando a capacidade dessas técnicas de se adaptarem a cenários eletroquímicos diversos. A variedade de algoritmos de aprendizado de máquina empregados, como ANNs, SVM, Naive Bayes, Random Forest, CNNs e DNNs, enfatiza a importância de escolher a abordagem mais adequada para cada aplicação específica.

Uma característica importante dos estudos analisados é a coleta e preparação de amostras de dados realizadas pelos próprios pesquisadores. Isso não apenas confere uma camada adicional de confiabilidade às pesquisas, mas também garante que os dados sejam pertinentes e adaptados aos seus objetivos. Por outro lado, embora a preparação autônoma das amostras pelos pesquisadores traga confiabilidade, ela também ressalta um desafio inerente na área: a ausência de bases de dados públicas. Esta lacuna pode ser um empecilho significativo para profissionais da computação ao tentar implementar algoritmos de aprendizado de máquina, dada a escassez de dados para diversas aplicações na área.

Muitos dos estudos alcançaram resultados sólidos, atingindo níveis significativos de precisão na identificação, quantificação e classificação de analitos em diferentes cenários eletroquímicos. Certos algoritmos, como SVM e XGBoost, destacaram-se como particularmente eficazes em várias aplicações. Ademais, várias pesquisas evidenciaram o potencial dessas técnicas na área da saúde, abrangendo desde a detecção de variantes do vírus SARS-CoV-2 até a previsão de níveis de biomarcadores em amostras de sangue.

Por fim, a utilização de técnicas de aprendizado de máquina tem o potencial de aprimorar significativamente a eficiência na detecção de eventos específicos em contextos eletroquímicos, resultando em economia de tempo e recursos. No entanto, é essencial considerar as limitações e desafios, como a ausência de bases de dados públicas, que podem afetar o desenvolvimento e aplicação de tais técnicas. Ainda assim, as abordagens inovadoras oferecem novos métodos para lidar com dados complexos, contribuindo para avanços em inúmeras aplicações, desde questões ambientais até a área da saúde. 

Como trabalhos futuros, podem ser analisados outros analitos utilizando aprendizagem de máquina para a detecção de substâncias químicas específicas em determinada amostra.

\bibliographystyle{sbc}
\bibliography{sbc-template}

\end{document}